\documentclass[runningheads]{llncs}
\usepackage{graphicx}
\usepackage{amsmath}
\usepackage{amssymb}
\usepackage{multirow}
\usepackage{listings}
\usepackage{makecell}
\usepackage{hyperref}

\begin{document}
\title{Valid Text-to-SQL Generation with Unification-based DeepStochLog}
\author{Ying Jiao\inst{1}\orcidID{0009-0009-2279-7691} \and
Luc De Raedt\inst{1, 2}\orcidID{0000-0002-6860-6303} \and \newline
Giuseppe Marra\inst{1}\orcidID{0000-0001-5940-9562}}
\authorrunning{Y. Jiao et al.}
\institute{KU Leuven, Dept. of Computer Science; Leuven.AI, B-3000 Leuven, Belgium
\email{\{firstname.lastname\}@kuleuven.be}\\
 \and
AASS, Örebro University, Sweden}
\maketitle              % typeset the header of the contribution
\begin{abstract}
  Large language models have been used to translate natural language questions to SQL queries. Without hard constraints on syntax and database schema, they occasionally produce invalid queries that are not executable. These failures limit the usage of these systems in real-life scenarios. We propose a neurosymbolic framework that imposes SQL syntax and schema constraints with unification-based definite clause grammars and thus guarantees the generation of valid queries. Our framework also builds a bi-directional interface to language models to leverage their natural language understanding abilities. The evaluation results on a subset of SQL grammars show that all our output queries are valid. This work is the first step towards extending language models with unification-based grammars. We demonstrate this extension enhances the validity, execution accuracy, and ground truth alignment of the underlying language model by a large margin. Our code is available at \url{https://github.com/ML-KULeuven/deepstochlog-lm}.

\end{abstract} 

\section{Introduction}

\begin{table}
  \caption{A comparison of our framework and the existing approaches. Syntax and schema information suggests the guidance of syntax and schema rules. Validity guarantee underscores the assurance that the output SQL queries are always executable. Learning and inference show if the methods can be applied during the learning and inference stages.  }
  \label{tab:relatedwork}
  \resizebox{\linewidth}{!}{\begin{tabular}{c|c|c|c|c|c}
    \hline
    &Syntax&Schema&Validity&\multirow{2}{*}{Learning}&\multirow{2}{*}{Inference}\\ 
    &information&information&guarantee&&\\
    \hline
    Neural-based : \cite{gao2023text}, \cite{pourreza2024din} & & & &-&-\\
    Sketch-based : \cite{xu2017sqlnet}, \cite{yu2018typesql}, \cite{yu2018syntaxsqlnet}, \cite{choi2021ryansql} & \checkmark & & &\checkmark&\checkmark\\
    Grammar-based : \cite{yin2018tranx}, \cite{guo2019towards}, \cite{lin2019grammar}, \cite{wang2020rat}  & \checkmark & & &\checkmark & \checkmark\\
    Constraint-based : \cite{scholak2021picard},  \cite{li2023graphix}, \cite{poesia2022synchromesh} & \checkmark & \checkmark & & & \checkmark\\
    Execution-guided : \cite{wang2018robust}, \cite{suhr2020exploring}, \cite{li2023resdsql}, \cite{dong2023c3} & \checkmark & \checkmark & & & \checkmark\\
   \textbf{Ours} & \checkmark & \checkmark &\checkmark & \checkmark &\checkmark\\ 
  \hline
\end{tabular}}
\end{table}

The text-to-SQL task is to map natural language sentences to SQL queries given database schema. It provides a natural language interface to empower users regardless of their technical background to access and derive value from vast relational databases. This task also plays a central role in emerging retrieval-augmented agents \cite{lewis2020retrieval,Chase_LangChain_2022,Liu_LlamaIndex_2022} for various applications, such as question answering, personal assistance, and intelligent customer service. While most existing studies focus on the accuracy of queries only, we emphasize the importance of validity, ensuring that they are executable. Invalid queries that fail to execute potentially introduce vulnerabilities to automatic agents.

Deep learning models, from early recurrent ones \cite{xu2017sqlnet,lin2019grammar} to recent large language models \cite{gao2023text,pourreza2024din}, have been successful in text-to-SQL. Though often effective, they can produce queries that violate SQL syntax and schema. Therefore, sketch- \cite{xu2017sqlnet,yu2018typesql,yu2018syntaxsqlnet,choi2021ryansql} and grammar-based approaches \cite{yin2018tranx,guo2019towards,lin2019grammar,wang2020rat} guided by context-free grammars have been proposed to avoid syntax errors. This idea is extended by constraint-based \cite{scholak2021picard,li2023graphix,poesia2022synchromesh} and execution-guided methods \cite{wang2018robust,suhr2020exploring,li2023resdsql,dong2023c3} by adding schema information. They filter errors at inference time but cannot be used at learning time. These methods cannot ensure the production of valid outputs as they can exit without finding a valid query in their search space. Table \ref{tab:relatedwork} summarizes the properties of the studies mentioned above.   

We present a neurosymbolic framework for text-to-SQL with a validity guarantee. Our framework uses DeepStochLog \cite{winters2022deepstochlog}, a sequence-based neural stochastic logic programming method, as a backbone. DeepStochLog introduces neural definite clause grammars (NDCGs) which integrate stochastic definite clause grammars and neural networks. Unlike grammars employed in previous works, the unification-based, Turing-complete definite clause grammars we use can represent any syntax and schema knowledge. Our generated queries are guaranteed to have no syntax or schema errors and are always valid. To apply DeepStochLog to the text-to-SQL task, we define LM definite clause grammars (LMDCGs), an extension of NDCGs for language models. They help harness the powerful language understanding capabilities of language models and handle dynamic variable domains. In experiments, we demonstrate the effectiveness of our approach in generating valid queries on a subset of SQL grammars. Our framework also substantially improves the alignment with ground truth queries and the execution accuracy of the underlying language model. In summary, our contributions are as follows:

\begin{itemize}
  \item We propose a neurosymbolic framework for text-to-SQL. To the best of our knowledge, we are the first to guarantee the production of valid queries with neural unification-based grammars.
  \item We introduce LMDCGs, an extension of DeepStochLog that integrates language models. 
  \item We empirically show that our neurosymbolic framework significantly improves the validity, ground truth alignment, and execution accuracy of the encapsulated language model. We surpass state-of-the-art text-to-SQL approaches in terms of validity. 
  \item We show the text-to-SQL task as a challenge and benchmark for neurosymbolic systems. 
\end{itemize}

\section{Problem Formulation}

Given a natural language sentence $nl$ and the schema $S$ of a database $db$, the text-to-SQL task translates $nl$ into an SQL query $q$. $S$ includes: 1) a set of tables $T = \{t_1, ..., t_N\}$ of size N, and 2) a set of columns $C = \{c_1^{1}, ..., c_{n_1}^{1}, ..., c_1^{N}, ..., c_{n_N}^{N}\}$ linked to the tables, where $n_i$ represents the number of columns in table $t_i$. 
% The total number of columns $M$ in $db$ is defined as $\sum_{i=1}^{N} n_i$.

\begin{figure} 
  \centering
  \includegraphics[width=\linewidth]{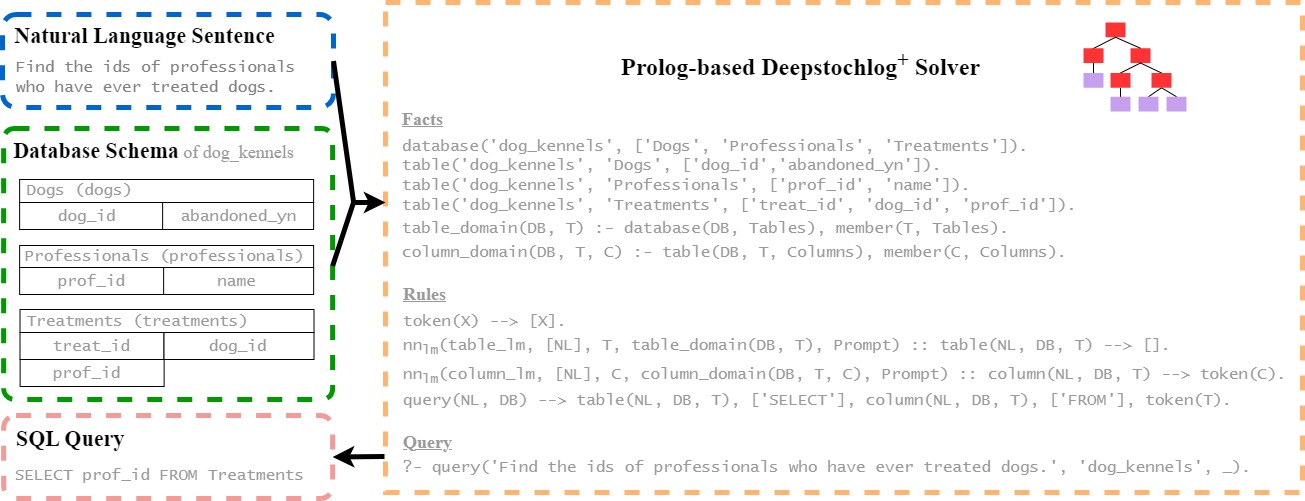}
  \caption{Illustration of a text-to-SQL instance solved by our framework (basic grammar is used for brevity). Given the inputs, the system maximizes the probability of the ground truth SQL query when it is known and produces the most probable query when the target is unknown. The first LMDCG rule $nn_{lm}$ in the logic program prompts the language model $table\_lm$ and gets a probability distribution over the three tables in the dog\_kennels database. Similarly, the second one prompts $column\_lm$ and gets a probability distribution over the columns in a given table. The inference steps are shown in Fig. \ref{fig:inference}.}  %"1.0:: " is omitted for rules with probability 1. }
  \label{fig:workflow}
\end{figure}

Fig. \ref{fig:workflow} demonstrates the workflow of our framework. We aim to generate a valid and correct $q$ that can retrieve the right answers to $nl$ from $db$ without runtime errors.

\section{Preliminaries} \label{sec:preliminary}

This section provides essential background information on grammar and DeepStochLog. We refer to the original DeepStochLog paper \cite{winters2022deepstochlog} for more details. 

\textbf{Context-free grammars (CFGs)} define a set of rewriting rules of the form $V \rightarrow W_1, ..., W_n$, where $V$ is a non-terminal and $W_i$ is either a terminal or a non-terminal. \textbf{Definite clause grammars (DCGs)} are a popular logic-programming-based extension of CFGs that can be executed as Prolog programs \cite{pereira1980definite}. They are unification-based and can encode context-sensitive languages. DCGs replace the non-terminals in CFGs by logical atoms $a(l_1, .., l_n)$ with a predicate $a$ and $n$ terms $l_i$. A term is a constant, a logical variable, or a structured term $f(l_1, ..., l_k)$ where $f$ is a functor. DCG rules take the format $nt\rightarrow  g_1, ..., g_n$, where $nt$ is an atom, a goal $g_1, ..., g_n$ is a sequence and $g_i$ is an atom or a list of terminals and logical variables. \textbf{Stochastic definite clause grammars (SDCGs)} formed as $p_i:: nt\rightarrow  g_1, ..., g_n$ add probabilities $p_i$ to DCG rules. They define a probability distribution over possible parses of a sequence and allow the most likely parse to be determined. SDCGs require the probabilities of the rules with the same non-terminal predicate to sum to 1.

DeepStochLog extends SDCGs to \textbf{neural definite clause grammars (NDCGs)} that integrate neural networks. An NDCG rule is defined as $nn( m,[I_1, ... , \newline I_X],[O_1, ... ,O_Y],[D_1, ... , D_Y]) :: nt \rightarrow  g_1, ... , g_n$. The $nn$ predicate denotes a neural network $m$ that takes input variables $I_1, ..., I_X$ and outputs a probability distribution over variables $O_1, ... ,O_Y$ with domains $D_1, ... ,D_Y$. For instance, $nn( table\_nn, [\textnormal{"Find ... dogs."}], T, [\textnormal{"Dogs", "Professionals", "Treatments"}]):: \newline table(\textnormal{"Find ... dogs."}) \rightarrow [T]$ represents a neural network $table\_nn$ that takes a natural language sentence "Find the ids of professionals who have ever treated dogs" as input and outputs a probability distribution over the table domain of "Dogs", "Professionals", "Treatments".

\begin{figure} 
  \centering
  \includegraphics[width=\linewidth]{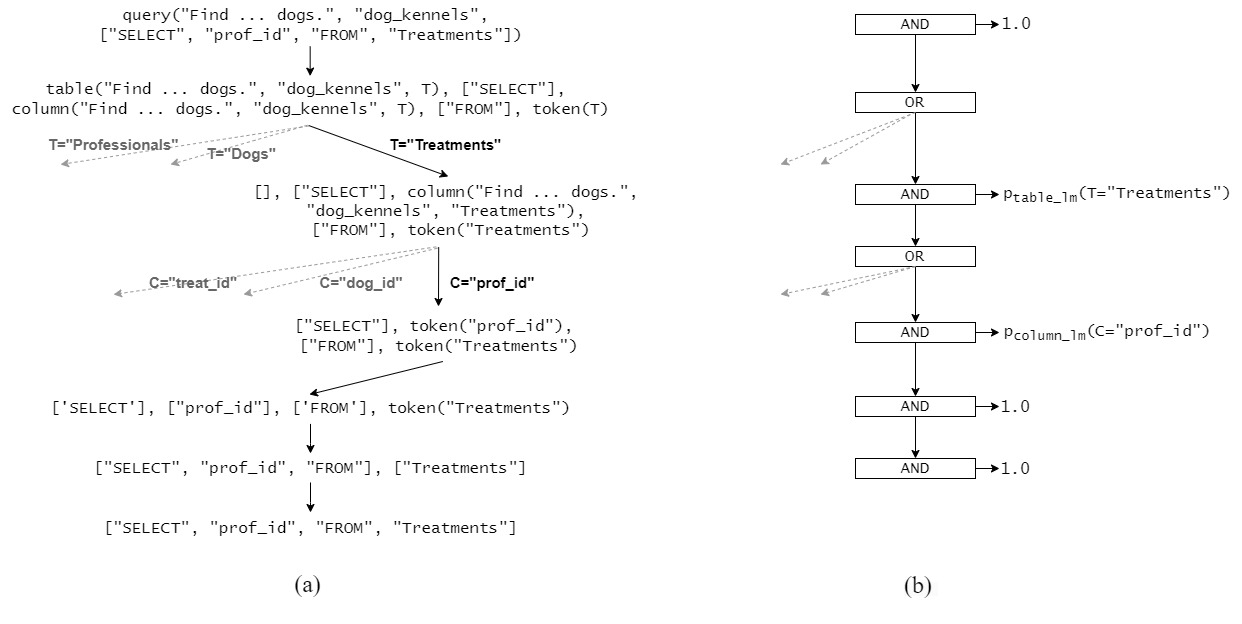}
  \caption{Inference steps on the text-to-SQL instance in Fig. \ref{fig:workflow}. (a) The SLD tree for $derives$($query$("Find the ids of professionals who have ever treated dogs.", "dog\_kennels", ["SELECT", "prof\_id", "FROM", "Treatments"])). Thanks to unification, the branches of the wrong table and column substitutions will fail. Failing branches are in grey.  (b) The corresponding AND-OR circuit. The probabilities of failing branches are not considered. }
  \label{fig:inference}
\end{figure}

Inference in DeepStochLog (see Section \ref{sec:workflow}) computes the probability of a logical goal given an input sequence and a DCG, using probabilities computed by neural networks. The set of parses of the input sequence is translated first into a logical proof tree (e.g. Fig. \ref{fig:inference} (a)), which is then turned into a computational graph (e.g. Fig. \ref{fig:inference} (b)) that computes the likelihood of the goal. In particular,  the probability $P_G(derives(G, T))$ is computed, where $G$ is a logical goal (e.g. the starting symbol) and $T$ is a sequence to parse. Logical inference uses resolution to find all derivations $d(G\theta)$ that produce $T$ with an answer substitution $\theta$. The resolution process is then translated into an AND-OR circuit. Probability inference calculates $P_G(derives(G, T))=\sum_{d(G\theta)=T}P((G\theta))=\sum_{d(G\theta)=T}\prod_{r_i \in d(G\theta)}p_i^{k_i}$, where $p_i$ is the probability of rule $r_i$ used for $k_i$ times in a derivation. This computation equals a bottom-up evaluation of the AND-OR circuit where the logical circuit is compiled to an arithmetic circuit with the $(+, \times)$ semiring \cite{kimmig2011algebraic}. Similarly, the most probable derivation for $G$ can be identified with the $(\max, \times)$ semiring. 

Learning in DeepStochLog (see Section \ref{sec:workflow})  is cast into the maximization of the likelihood of the input sequences.  It is defined as
\begin{equation}
\label{eq:learning}
    \min_{p} \sum\nolimits_{(G_i\theta_i, T_i, t_i) \in \mathcal{D}} \mathcal{L}(P_G(derives(G_i\theta_i,T_i);p), t_i)
\end{equation}
where $p$ is a vector of rule probabilities, $\mathcal{D}$ is a dataset of triples $\{G_i\theta_i, T_i, t_i\}$, $t_i$ is a target probability and $\mathcal{L}$ is a differentiable loss function. This learning problem is solved with standard gradient descent techniques like the Adam optimizer \cite{kingma2014adam}. The gradients of $\mathcal{L}$ w.r.t $p$ can be computed automatically and backpropagated seamlessly to train the internal parameters of neural networks.

\section{Methodology}

We propose a neurosymbolic framework for text-to-SQL based on LM definite clause grammars (LMDCGs). LMDCGs are an extension of neural definite clause grammars (NDCGs) that integrate stochastic definite clause grammars (SDCGs) with language models. In Section \ref{sec:workflow}, we show the workflow of our framework with the text-to-SQL instance in Fig. \ref{fig:workflow}. We define LMDCGs and illustrate the bi-directional interface to language models they provide in Section \ref{sec:lmrule}. Lastly, we demonstrate the advantage of our definite clause grammars (DCGs) over the rules in the previous sketch- and grammar-based approaches in Section \ref{sec:dcgcfg}. 

\subsection{Our Workflow} \label{sec:workflow}
 
Our logic program has three parts: facts, rules, and a Prolog query. 

\paragraph{Facts} The facts represent associations between the database, tables, and columns. They are automatically generated from the database schema to define the domain of possible table and column variable substitutions. 
For example, in Fig. \ref{fig:workflow}, the fact with $database$ predicate describes the three tables in the database "dog\_kennels". The $table$ predicate describes the columns in each table. The $table\_domain$ and $column\_domain$ predicates retrieve the tables in dog\_kennels and the columns in a given table respectively.

\paragraph{Rules} The rules encoding DCGs will be used to find answers to $query(nl, db, Q)$, where $Q$ is the ground truth SQL query. Each rule can be assigned a probability. In our task, language models determine probabilities of LMDCG rules using the predicate $nn\_lm$ (see Section \ref{sec:lmrule}). Fig. \ref{fig:inference} (a) shows how the rules are applied to parse the ground truth SQL query ["SELECT", "prof\_id", "FROM", "Treatments"] step by step. As in Fig. \ref{fig:inference} (b), $table\_lm$ and $column\_lm$ determine the probability of the table and column grammar branches respectively. The $query$ and $token$ rules are deterministic, i.e. purely logical, with a probability of 1.0.

\paragraph{Prolog Query} The Prolog query\footnote{Notice that the Prolog query is the logical goal to be proved, which differs from the SQL query to generate.} $query(nl, db, Q)$ defines the output of our framework. During training, $Q$ is known and the system outputs the probability of producing $Q$ given natural language sentence $nl$ and database $db$. To this end, DeepStochLog inference with the $(+, \times)$ semiring is used  (see Section \ref{sec:preliminary}). For example, in Fig. \ref{fig:inference}, given the sentence "Find ... dogs" and the database dog\_kennel, the probability of the ground truth SQL query 
\begin{align*}
P(derives(query(\textnormal{"Find ... dogs.", "dog\_}  \textnormal{kennels", ["SELECT", "prof\_id",} 
\end{align*}
\vspace{-9mm}
\begin{align*}
\textnormal{"FROM", "Treatments"]}))) = 1.0 \times ( 0+0 + P_{table\_lm}(\textnormal{T="Treatments"}) \times  
\end{align*}
\vspace{-9mm}
\begin{align*}
(0+0 + P_{column\_lm} (\textnormal{C="prof\_id"}) \times 1.0 \times 1.0))
\end{align*}  The learning process maximizes the probability of this query. Since DeepStochLog produces end-to-end differentiable inference graphs (i.e. AND/OR circuit in Fig. \ref{fig:inference} (b)), this can be easily achieved by standard backpropagation and stochastic gradient descent. 
During evaluation, the  $Q$ is unknown. The system outputs the most likely SQL query given $nl$ and $db$. To this end, DeepStochLog inference with the $(\max, \times)$ semiring is used.

\subsection{LMDCGs} \label{sec:lmrule}

We define an LMDCG rule as:
\begin{displaymath}
    nn_{lm}( lm, [NL], O_Y, D_Y, Prompt) :: nt \rightarrow  g_1, ... , g_n
\end{displaymath}
where $lm$ is a language model, $NL$ is a natural language sentence, $O_Y$ is an output variable with domain $D_Y=[y_1, ..., y_n]$, and $Prompt$ is a constant sentence. 

LMDCGs communicate with the underlying language model bi-directionally: 1) by constructing the input to execute them; and 2) by re-normalizing their output to build probabilities for SDCGs. Since the probabilities produced by language models are used during inference, we can backpropagate gradients seamlessly to language models and fine-tune them.

\paragraph{Language Model Input Construction} In the text-to-SQL task, the language model input is designed as the concatenation of 1) $NL$, 2) the possible substitutions of $O_Y$ with their indexes in $D_Y$, i.e. "Answer $i$ for $y_i$", and 3) $Prompt=$ "the answer should be Answer". The substitutions of $O_Y$ in $D_Y$ can be tokenized into several parts with language models, for example, the column name "abandoned\_yn". We require the language models to output indexes instead of the substitutions to make them treat every substitution as a whole. For example, in Fig. \ref{fig:workflow}, given the natural language sentence "Find ... dogs.", the table domain ["Dogs", "Professionals", "Treatments"], and the rule "$nn_{lm}(table\_lm, [NL], T, table\_domain$ $(DB, T), Prompt):: table(NL, DB, T) \rightarrow [].$", the input to the $table\_lm$ is "Find ... dogs. Answer 1 for Dogs, Answer 2 for Professionals, Answer 3 for Treatments, the answer should be Answer ". $table\_lm$ takes this input and outputs a logit for every token in its vocabulary.

\paragraph{Language Model Output Normalization} To get the probability distribution over $D_Y$, for language models with a decoder, we extract the logits for indexes $i$ from the decoder outputs and renormalize them with the softmax function. For the encoder-only language models, we apply a linear layer on top of them. For example, in Fig. \ref{fig:workflow},  the logits for the token "1", "2" and "3" are extracted and the softmax distribution could be 0.2, 0.2, and 0.6. Thus, the LMDCG rule represents a set of grammar branches: $0.2:: table(\textnormal{"Find ... dogs.", "dog\_kennels", "Dogs"})\rightarrow [];  0.2:: table(\textnormal{"Find ...   dogs.",} \textnormal{"dog\_kennels", "Professionals"})\rightarrow [];  0.6:: table( \newline \textnormal{"Find ...} \textnormal{dogs.", }  \textnormal{"dog\_kennels",}  \textnormal{ "Treatments"})\rightarrow []$. Here, we apply the empty production. The rule produces an empty sequence but provides substitutions of the table variable $T$, which helps determine the column domain in the column rule through the Prolog unification mechanism. Unlike the autoregressive models, the empty production allows us to deal with tables before columns which better fits human intuitions. 

\subsection{DCGs v.s. Previous Rules} \label{sec:dcgcfg}

The rules in previous sketch- \cite{xu2017sqlnet,yu2018typesql,yu2018syntaxsqlnet,choi2021ryansql} and grammar-based approaches \cite{yin2018tranx,guo2019towards,lin2019grammar,wang2020rat} do not explicitly represent relationships between tables and columns. When generating basic SQL queries, they are equivalent to CFGs in Example \ref{ex:cfgdcg} a). Considering the dog\_kennels database in Fig. \ref{fig:workflow}, these approaches can overgenerate and lead to invalid queries mismatching tables and columns, for example, "SELECT treat\_id FROM Dogs". 

The example DCG avoids this error and encodes the associations of tables and columns by bounding the column domain to columns in a specific table. We demonstrate that DCGs can guarantee the correctness of syntax and the faithfulness to schema and thus guarantee the validity of outputs with this basic SQL generation scenario. As DCGs are Turing-complete, they can be generalized to express sophisticated SQL syntax and semantic constraints and produce valid, advanced SQL queries.

\begin{example}
\label{ex:cfgdcg}

An equivalent CFG of rules in previous works a) and a DCG b).

\vspace{-3mm}

\begin{minipage}[t]{0.8\textwidth}
\begin{align*}
    a)&\ T \rightarrow  \textnormal{"Dogs", "Professionals", "Treatments"}\\
    &\ C \rightarrow \textnormal{"dog\_id", "abandoned\_yn", "prof\_id", "name", "treat\_id"}\\
    &\ Q \rightarrow \textnormal{"SELECT"}\  C\ \textnormal{"FROM"}\ T \\
\end{align*}
\end{minipage}

\vspace{-6mm}

\begin{minipage}[t]{0.8\textwidth}
\begin{align*}
    b)&\ table(\textnormal{"Dogs"}) \rightarrow \textnormal{"Dogs"}\\
    &\ table(\textnormal{"Professionals"}) \rightarrow \textnormal{"Professionals"}\\
    &\ table(\textnormal{"Treatments")} \rightarrow \textnormal{"Treatments"}\\
    &\ column(\textnormal{"Dogs"}) \rightarrow \textnormal{"dog\_id", "abandoned\_yn"}\\
    &\ column(\textnormal{"Professionals"}) \rightarrow \textnormal{"prof\_id", "name"}\\
    &\ column(\textnormal{"Treatments"}) \rightarrow \textnormal{"treat\_id", "dog\_id", "prof\_id"}\\
    &\ Q \rightarrow \textnormal{"SELECT"}\  column(T)\ \textnormal{"FROM"}\ table(T)
\end{align*}
\end{minipage}
  
\end{example}

\section{Experiments}

\subsection{Research Questions}

Our experiments aim to address the following questions:
\begin{itemize}
 \item[\textbf{Q1}] Which language model produces more correct queries?

 \item[\textbf{Q2}] Does our framework ensure validity? How does it compare to other text-to-SQL approaches?
 
 \item[\textbf{Q3}] Does our framework improve exact matching and execution accuracy?

\end{itemize}

\subsection{Tasks} 

\paragraph{Task 1 (Q1)} We explore which language model should we encapsulate to achieve better performance. The language models in our framework can be classified into two types. Type 1 produces probability distributions over table and column grammar branches with dynamic domains. For type 1, we use T5-small models with an encoder-decoder structure. T5 \cite{raffel2020exploring} is popular in addressing the text-to-SQL task \cite{shaw2021compositional,scholak2021picard}. The vocabulary of its decoder allows us to handle the domains of the table and the column that vary with the database. Type 2 provides probability distributions over grammar branches defined by SQL syntax with fixed output domains. For example, the selection branches between "SELECT" and "FROM" could be "*", "COUNT(*)", a column with an aggregation function, etc. 

In task 1, we compare T5-small with Bert-base \cite{devlin2019bert} plus a linear layer for type 2 using a grammar for the SELECT clause (in Appendix \ref{sec:program1}). Setting 1 uses two T5-small models and one Bert-base for the table, column, and selection branch respectively. Setting 2 uses three T5-small models. The training and evaluation details for task 1 are in Appendix \ref{sec:traineval1}. 

\paragraph{Task 2 (Q2, Q3)} We evaluate our system with a recursion-free grammar covering the SELECT, WHERE, GROUP BY, ORDER BY clauses and EXCEPT between two simple SELECT clauses. The recursive cases are left for future studies. The exact inference in Section \ref{sec:preliminary} is inefficient for the extended grammar. We use the greedy inference that takes the most likely grammar branch. Appendix \ref{sec:task2} describes more training and evaluation details, the grammar, and the language models employed. 

We compare our framework with the following baselines:
\vspace{-2mm}
\begin{itemize}
   \item Neural-based:
   \begin{itemize}
     \item T5-small: fine-tuned T5-small that treats text-to-SQL as sequence-to-sequence generation.
     \item DAIL-SQL \cite{gao2023text} and DIN-SQL \cite{pourreza2024din}: GPT-4 \cite{achiam2023gpt} under few-shot prompting.
   \end{itemize}
   \vspace{2mm}
    \item Grammar-based:
    \begin{itemize}
     \item T5-small + CFGs: an ablation of our framework which does not consider the relations between table and column. It simulates the rules in previous sketch- \cite{xu2017sqlnet,yu2018typesql,yu2018syntaxsqlnet,choi2021ryansql} and grammar-based approaches \cite{yin2018tranx,guo2019towards,lin2019grammar,wang2020rat}.
    \end{itemize}
    \vspace{2mm}
    \item Constraint-based:
    \begin{itemize}
     \item Graphix-T5 \cite{li2023graphix}: T5-3B augmented with graph-aware layers and constrained decoding PICARD \cite{scholak2021picard}.
    \end{itemize}
    \vspace{2mm}
    \item Execution-guided:
    \begin{itemize}
     \item C3 \cite{dong2023c3}: zero-shot ChatGPT. The final output is selected based on execution results. 
    \end{itemize}
 \end{itemize}
More information on the methodology of baselines is in Section \ref{sec:relatedwork}. Their implementation details are in Appendix \ref{sec:baselines}

Following \cite{yu2018spider}, we consider evaluation metrics: exact matching that compares the predicted and the ground truth query, and the execution accuracy that compares their execution results. We also report validity that checks whether the predicated queries are executable.

\paragraph{Data} We extract samples that satisfy the scope of our grammar from Spider \cite{yu2018spider}, a large-scale complex and cross-domain benchmark dataset for text-to-SQL. All models are evaluated on the instances extracted from Spider's development division. In task 1, we employ 384 training samples and 59 evaluation samples. Task 2 uses 2106 training samples and 258 evaluation samples.

\subsection{Results}

\paragraph{Q1} For task 1 setting 1 that uses Bert-base for the selection branch, the percentage of outputs with correct execution results is limited to 61\%. Its results always start with "SELECT COUNT(*)". This is caused by the unbalanced training data shown in Table \ref{tab:selection}. Setting 2 using T5-small for the selection branch is less affected by the biased data. 93\% outputs lead to the right results. Since the training data for grammar branches is often unbalanced, we encapsulate T5-small for all neural components in task 2 experiments.  

%It equals the proportion of samples with "COUNT(*)" in all evaluation samples. 

\begin{table}
  \caption{ Selection branch statistics of task 1 training data. col. stands for column.}
  \label{tab:selection}
  \resizebox{\linewidth}{!}{\begin{tabular}{c|c|c|c|c|c|c|c}
    \hline
    *&COUNT(*)&col.&COUNT(col.)&SUM(col.)
    &AVG(col.)&MIN(col.)&MAX(col.)\\
    \hline
   2.1\% & 55.5\% & 14.9\% & 3.7\% & 7.3\% & 12.1\% & 0.5\% & 3.9\%\\
  \hline
\end{tabular}}
\end{table}

\paragraph{Q2} Table \ref{tab:results} compares our framework with state-of-the-art text-to-SQL approaches. Our DeepStochLog with LMDCGs is able to guarantee validity in 100\% of the test queries. 
%shows our framework guarantees validity. It demonstrates the effectiveness of our neural unification-based grammars. 
We ensure faithfulness to both SQL syntax and database schema.
In comparison, neural-based methods (T5-small, DAIL-SQL, and DIN-SQL) without hard schema constraints can produce non-valid queries by using identifiers not defined in the schema. T5-small + CFGs can mismatch tables and columns due to the lack of constraints on their relations. Graphix with constrained decoding can exit without finding a valid query and output incomplete results. C3 with execution-guided decoding also produces 100\% valid queries for the extracted evaluation examples. However, execution is not always feasible in application scenarios. Table \ref{tab:invalid} shows examples of common errors made by the baselines.  

%Our generated SQL queries are 100\% executable.

\begin{table*} \centering
  \caption{Comparison with state-of-the-art models for text-to-SQL on the selected subset of Spider. (*) Execution-based methods are not applicable in real settings as they need to execute the query during generation. Params. means parameters. }
  \label{tab:results}
  \resizebox{\linewidth}{!}{\begin{tabular}{c|c|c|c|c}
    \hline
    & & \multirow{2}{*}{Validity\%} & Exact  & Execution   \\
    &&& Matching \% & Accuracy\% \\
    \hline
    \multirow{3}{*}{\makecell[c]{Smaller Models \\ (Millions Params.)}}&T5-small & 53.9 & 41.1 & 41.1 \\
    &T5-small+CFGs & 88.8 & 67.1 & 70.9 \\
    &Ours (T5-small+DCGs) & \textbf{100.0} & 75.6 & 77.9  \\
    \hline
     \multirow{3}{*}{\makecell[c]{Larger Models \\ (B/Trillions Params.)}}&DAIL-SQL (GPT-4) & 99.2 & 88.8 & 89.9  \\
    &DIN-SQL (GPT-4) & 99.2 & 78.7 & 90.7  \\
    &Graphix-T5 (T5-3B+PICARD) & 99.6 & \textbf{91.9} & \textbf{91.9}  \\
    \hline
    Execution Required & C3 (ChatGPT+Execution)  & \textbf{100.0}* & 80.6* & 85.3* \\
  \hline
\end{tabular}}
\end{table*}

\begin{table*}
  \caption{Examples of invalid outputs from baseline models.}
  \label{tab:invalid}
  \resizebox{\linewidth}{!}{\begin{tabular}{c|c|c}
    \hline
    &Example& Error \\
    \hline
    \multirow{2}{*}{T5-small} & SELECT Name FROM country  & Invent identifiers \\
    & WHERE Independence > 1950 & Independence not in schema \\
    \hline
    \multirow{2}{*}{T5-small+CFGs}& SELECT COUNT(*) FROM Has\_Pet  & Mismatch table and column, \\
    & WHERE weight > 10 & weight not in Has\_Pet \\
    \hline
    \multirow{2}{*}{DAIL-SQL}& SELECT Paragraph\_Details FROM Paragraphs & Invent identifiers \\
    & WHERE Paragraph\_Text LIKE '\%Korea\%' & Paragraph\_Details not in schema \\
    \hline
    \multirow{2}{*}{DIN-SQL}& SELECT T1.first\_name, ... & Invent identifiers \\
    &ORDER BY T2.rank\_points DESC LIMIT 1& rank\_points not in schema \\
    \hline
    Graphix & select & Incomplete query \\
  \hline
\end{tabular}}
\end{table*}

\paragraph{Q3} The performance on exact matching and execution accuracy (see Table \ref{tab:results}) also suggests the effectiveness of integrating unification-based grammars with language models. Compared to the vanilla T5-small model, our framework that extends T5-small with unification-based grammars improves the exact matching and execution accuracy by a large margin. We also outperform T5-small + CFGs due to the bounding of column domains to corresponding tables. However, T5-small, with 60 million parameters, limits the capability of our framework to produce correct SQL queries. As shown in Table \ref{tab:incorrect}, our outputs with incorrect execution results are caused by misunderstanding user intentions and linking the natural language sentences to the wrong SQL components. The results of DAIL-SQL, DIN-SQL, Graphix, and C3 indicate that this problem could be alleviated by accessing larger-scaled models with billions or trillions of parameters and by employing graph-aware layers that model relations better.

\begin{table*}
  \caption{Examples of incorrect outputs from our framework. Pred. refers to predication. }
  \label{tab:incorrect}
  \resizebox{\linewidth}{!}{\begin{tabular}{c|c|c}
    \hline
    Mislinking Type&Example& Proportion \% \\
    \hline
    \multirow{3}{*}{Identifier(s)} & Find the number of distinct name of losers. & \multirow{3}{*}{65.0}\\
    \cline{2-2}
    & Pred.: SELECT COUNT( DISTINCT first\_name ) FROM players  &  \\
    \cline{2-2}
    & Gold: SELECT count(DISTINCT loser\_name) FROM matches &  \\
    \hline
    \multirow{3}{*}{Selection branch}& Count the number of dogs that went through a treatment. & \multirow{3}{*}{19.7}\\
    \cline{2-2}
    & Pred: SELECT COUNT(*) FROM Treatments & \\
    \cline{2-2}
    & Gold: SELECT count(DISTINCT dog\_id) FROM Treatments & \\
    \hline
    \multirow{4}{*}{Clause} & Count the number of high schoolers. & \multirow{4}{*}{12.0} \\
    \cline{2-2}
    & Pred: SELECT COUNT(*) FROM Highschooler & \\
    &  WHERE grade = 1 & \\
    \cline{2-2}
    & Gold: SELECT count(*) FROM Highschooler &  \\
    \hline
    \multirow{5}{*}{Operator} & Which cities do more than one employee under age 30  come from? & \multirow{5}{*}{3.3}\\
    \cline{2-2}
    & Pred: SELECT City FROM employee WHERE Age > 30  &  \\
    &  GROUP BY City HAVING COUNT(*) > 1 & \\
    \cline{2-2}
    & Gold: SELECT city FROM employee WHERE age  <  30 &  \\
    & GROUP BY city HAVING count(*)  >  1 & \\
  \hline
\end{tabular}}
\end{table*}

\section{Related Work} \label{sec:relatedwork}

\paragraph{Neural-based} The state-of-the-art text-to-SQL approaches are based on large language models (LLMs). They do not employ any hard constraints to guarantee valid outputs. DAIL-SQL \cite{gao2023text} designs the few-shot prompt for LLMs. It uses code question representation and selects examples based on both question and query. DIN-SQL \cite{pourreza2024din} decomposes the text-to-SQL task into sub-problems: schema linking, classification and building different prompts from each class, SQL generation, and self-correction. The first three steps are conducted by LLMs under the few-shot setting and the last one by instructing a LLM.

\paragraph{Sketch-based} Sketch-based methods \cite{xu2017sqlnet,yu2018typesql,yu2018syntaxsqlnet,choi2021ryansql} model text-to-SQL as a sequence-to-set problem considering the possible equivalent serialization of one query. They define a dependency graph of slots filled by independently trained neural components. The query synthesis process is viewed as an inference on the graph. Their sketches agree with SQL syntax but do not encode schema information like table and column relations. Therefore, the produced queries can be non-executable due to violations of schema. 

\paragraph{Grammar-based} Grammar-based methods \cite{yin2018tranx,wang2020rat,guo2019towards,lin2019grammar} introduce a sequence-to-action formalism of text-to-SQL that generates a derivation Abstract Syntax Tree \cite{yin2017syntactic} or a similar intermediate representation \cite{guo2019towards}. At each time step, they use context-free grammars to define the possible actions, and a trained neural model to predict the probability distribution over the action set. Similar to sketch-based approaches, these probabilistic grammar models cannot ensure the validity of their outputs as they do not include semantic constraints.

\paragraph{Constraint-based and Execution-guided} Constraint-based methods \cite{scholak2021picard,poesia2022synchromesh} accept only tokens that align with defined SQL syntax and semantic constraints during decoding. Graphix-T5 \cite{li2023graphix} applies constraint-based PICARD \cite{scholak2021picard} to prune erroneous tokens during its beam-search phase. It also enhances T5 with graphix layers to better model the relational structures in text-to-SQL. Execution-guided approaches \cite{wang2018robust,suhr2020exploring,li2023resdsql} execute queries and filter out faulty ones during generation. C3 \cite{dong2023c3} samples a set of SQL queries from zero-shot ChatGPT and votes for the most consistent one based on their execution results. The methods in these two categories are effective but can fail to generate valid outputs when the underlying models put low probabilities on valid predictions.

\section{Conclusion and Future Work}

 We introduce LM definite clause grammars (LMDCGs), an extension of DeepStochLog \cite{winters2022deepstochlog} that integrates language models with unification-based grammars to provide a validity guarantee to the text-to-SQL task. We evaluate our method on a subset of SQL syntax. The results suggest that this integration eliminates non-executable queries and significantly contributes to the alignment with ground truth queries and execution accuracy. 

Several limitations of our framework are interesting to explore in future studies. First, the prompt part of LMDCGs is not yet built dynamically to indicate the parsing states. Second, our system could be further enhanced by employing billion-level open-source large language models like Llama \cite{touvron2023llama}, Falcon \cite{falcon40b}, and Alpaca \cite{alpaca}. Lastly, scaling the current framework to larger grammars is not trivial. An interesting direction for speeding up the inference process would be searching for the k-best derivations.

\section*{Acknowledgments}
This project has received funding from the European Union’s Horizon Europe research and innovation programme under the Marie Skłodowska-Curie grant agreement No 101073307 and the Flemish Government (AI Research Program). Luc De Raedt is also supported by
the Wallenberg AI, Autonomous Systems and Software Program (WASP) funded by the Knut and Alice Wallenberg Foundation. We thank Thomas Winters for the helpful discussions. We also thank the anonymous reviewers for their valuable feedback.
\vspace{-3mm}
\begin{figure}
      \includegraphics[width=0.35\linewidth]{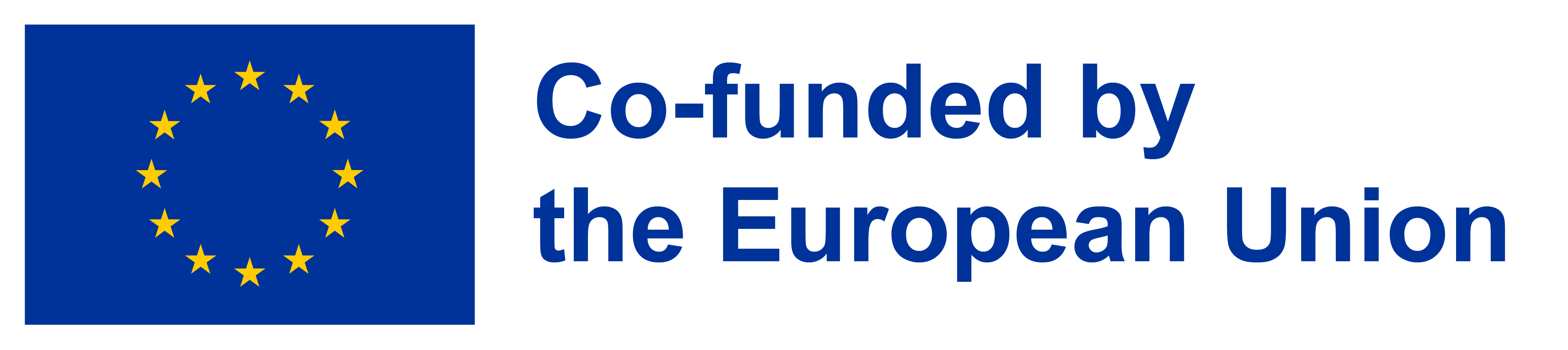}
\end{figure}

\bibliographystyle{splncs04}
\bibliography{ours}

\newpage
\appendix

\section{Task 1}

\subsection{Logic Program} \label{sec:program1}

Task 1 grammar is modeled as follows:

\vspace{-5mm}

\begin{figure} 
  \centering
  \includegraphics[width=\linewidth]{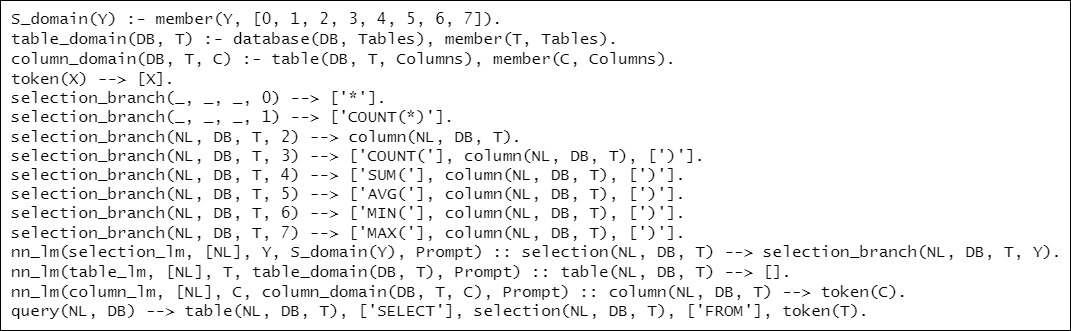}
\end{figure}

\vspace{-5mm}

\subsection{Training and Evaluation} \label{sec:traineval1}

We train settings 1 and 2 end-to-end for 7 epochs using the Adam optimizer \cite{kingma2014adam} with a batch size of 8 and a learning rate of $1e^{-3}$. For evaluation, we employ DeepStochLog inference with the $(\max, \times)$ semiring, i.e. the exact inference. Examples of $column\_lm$ and $selection\_lm$ inputs are listed in Table \ref{tab:ourprompt}. The inputs of $table\_lm$ have the same format as those of $column\_lm$.  

\paragraph{Pre- and post-processing} In pre-processing, we tokenize the ground truth SQL queries to sequences with list format. Table and column identifiers in the sequences are replaced by their semantic names \cite{li2023resdsql}. The semantics names provided in Spider \cite{yu2018spider} are closer to natural expressions, which facilitate the understanding of language models. This replacement is also conducted for the facts in logic programs. After generation, we restore the identifiers in the output sequence to their original names and join the sequence to get the SQL query. This pre- and post-progressing are also performed in task 2 experiments.

\section{Task 2} \label{sec:task2}

\subsection{Logic Program}

Task 2 grammar covers two types of queries: 1) single selection and 2) two selections connected by the set operator "EXCEPT". The grammar for single-selection queries covers the "SELECT", "WHERE", "GROUP BY", and "ORDER BY" clauses including "DISTINCT" and aggregation functions in the "SELECT" clause, "HAVING" in the "GROUP BY" clause, and "ASC / DESC" and "LIMIT" in the "ORDER BY" clause. "WHERE" and "HAVING" allow one condition. For type 2), the two selection clauses have the format "SELECT [column] FROM [table]". The columns in the two selection clauses are currently restricted to foreign keys linking the two tables. Task 2 grammar is modeled as follows:

\begin{figure} [t]
  \centering
  \includegraphics[width=\linewidth]{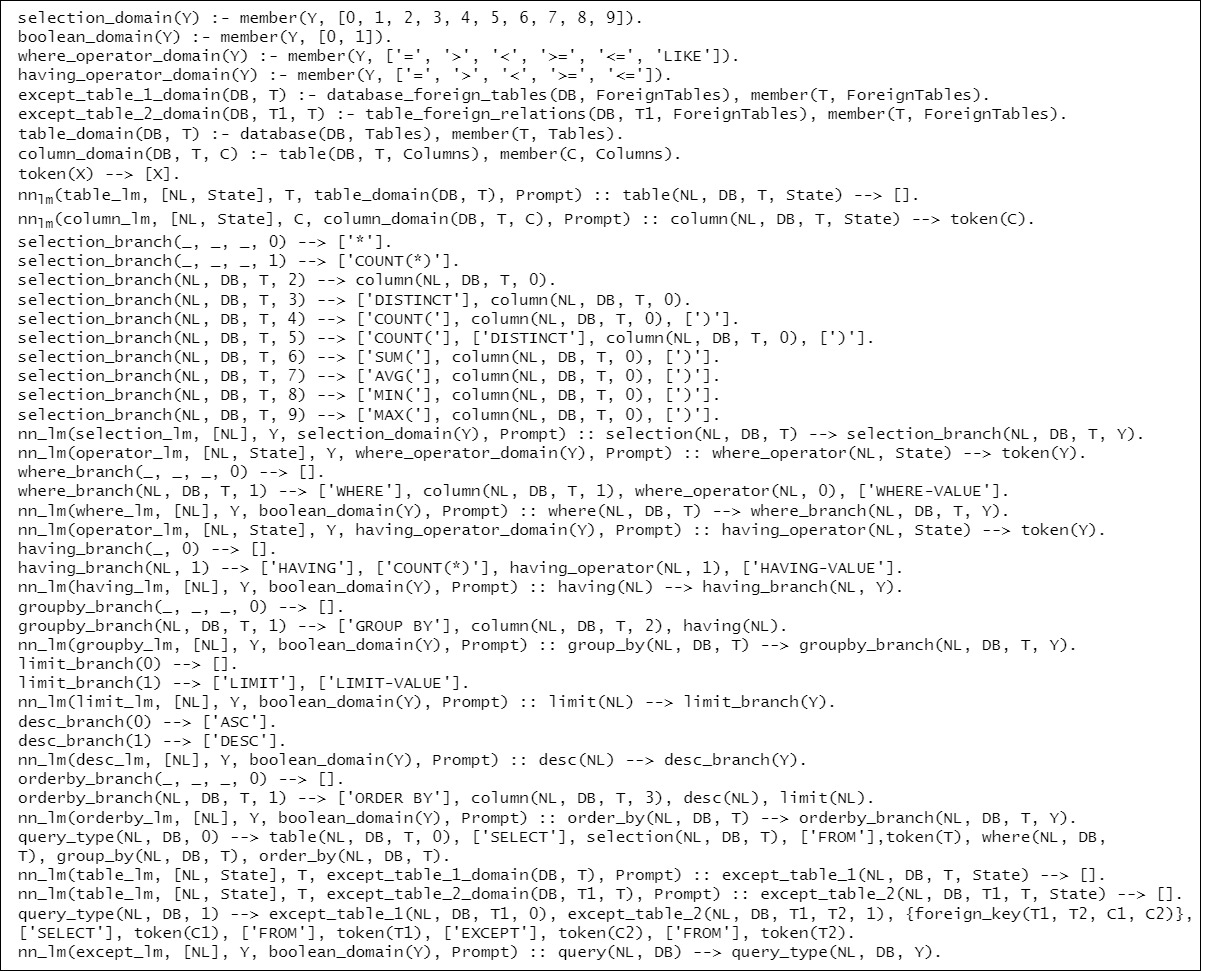}
\end{figure}

\subsection{Underlying Language Models}

We employ 11 fine-tuned T5-small models. All models used in this work are from Huggingface \cite{wolf2020transformers}.

$table\_lm$ and $column\_lm$ provide the probability distribution over the given table and column domain respectively. Their output domains vary with the database. The output domain of other language models is fixed. $except\_lm$, $where\_lm$, $groupby\_lm$, $having\_lm$, $order\_lm$, $desc\_lm$, $limit\_lm$ are used to produce the probability distribution on the existence of the corresponding SQL clause. $selection\_lm$ provides the probability distribution over 10 possible selection branches. $operator\_lm$ outputs the probability distribution over possible operators in "WHERE" and "HAVING" conditions. For the queries with two selection clauses connected by "EXCEPT", we call $table\_lm$ twice to get probability distributions over the possible substitutions for the table in each selection clause. The pair of columns in each selection clause are decided deterministically based on the pair of tables. Our current framework does not predict any value in SQL conditions. We assume the gold values are given. When we predict wrong conditions that cannot be mapped to the gold values, we assign the values to 1.

$table\_lm$, $column\_lm$, and $opertor\_lm$ can be used at different positions. $table\_lm$ can be called twice for the selection clause before "EXCEPT" and the one after "EXCEPT". $column\_lm$ is used for the column in the "SELECT", "WHERE", "GROUP BY" and "ORDER BY" clauses. $operator\_lm$ is used for operators in "WHERE" and "HAVING" conditions. We add states in their inputs to help them distinguish different cases. Examples of $column\_lm$ inputs are shown in Table \ref{tab:ourprompt} to showcase the inputs with states. We also include examples of $where\_lm$ and $selection\_lm$ inputs in Table \ref{tab:ourprompt}. $except\_lm$, $where\_lm$, $groupby\_lm$, $having\_lm$, $order\_lm$, $desc\_lm$, $limit\_lm$ shares a similar input format as $where\_lm$. 

\subsection{Training and Evaluation} 

In the text-to-SQL task, the grammar can always be written unambiguously, which leads to only one possible derivation for the ground-truth query $Q$. With the negative log-likelihood loss function and all positive samples in dataset $\mathcal{D}$ (target probability $t_i$=1.0),
\begin{displaymath}
\begin{aligned}
     (\ref{eq:learning}) &= \min_{p} \sum\nolimits_{(G_i\theta_i, Q_i) \in D} -log(\prod\nolimits_{r_j \in d(G_i\theta_i)=Q_i}p_j^{k_j}) \\
     &= \sum\nolimits_{(G_i\theta_i, Q_i) \in D} \sum\nolimits_{r_j \in d(G_i\theta_i)=Q_i} \min_{p_j} -k_jlogp_j
\end{aligned}
\end{displaymath}
As all the intermediate goals are observable given $Q$, the learning problem collapses to supervised training. 

We fine-tune the T5-small models using the Adam optimizer. Table \ref{tab:hyper} shows the number of fine-tuning epochs, the batch size, and the learning rate for each model. The hyper-parameters are determined with cross-validation. 

In evaluation, we perform the greedy inference to speed up the inference process. Instead of considering the re-normalized distributions obtained from the language models, the greedy inference takes the substitution with the largest probability.

\begin{table}
  \caption{Fine-tuning hyper-parameters of our T5-small models.}
  \label{tab:hyper}
  \resizebox{\linewidth}{!}{\begin{tabular}{c|c|c|c|c|c|c|c|c|c|c|c}
    \hline
    &table&column&except&where&group by
    &having&order by&desc&limit&selection&operator\\
    \hline
    Epochs&5&16&3&14&4&2&7&10&3&7&17\\ \hline
    \makecell[c]{Batch \\size}&64&32&64&32&64&64&64&32&64&64&64\\ \hline
    \makecell[c]{Learning \\rate}&$1e^{-3}$&$5e^{-4}$&$5e^{-4}$&$5e^{-4}$&$1e^{-3}$&$5e^{-4}$&$1e^{-3}$&$5e^{-4}$&$5e^{-4}$&$1e^{-3}$&$1e^{-3}$\\ \hline
\end{tabular}}
\end{table}

\section{Baselines} \label{sec:baselines}

We fine-tune the vanilla T5-small baseline for 10 epochs using the Adam optimizer, a batch size of 32, and a learning rate $1e^{-3}$. T5-small + CFGs shares the same language models with ours T5-small + DCGs except the $column\_lm$. Without table unification, the domain of $column\_lm$ used in T5-small + CFGs covers all the columns in a given database. We fine-tune a T5-small model for $column\_lm$ in T5-small + CFGs using the Adam optimizer for 13 epochs with a batch size of 32, and a learning rate $5e^{-4}$. Table \ref{tab:baselineprompt} shows the example inputs for T5-small and $column\_lm$ in T5-small + CFGs.

For the state-of-the-art models (DAIL-SQL \cite{gao2023text}, DIN-SQL \cite{pourreza2024din}, Graphix-T5 \cite{li2023graphix}, and C3 \cite{dong2023c3}), we extract their predictions on the samples in our evaluation set from their official results for the full Spider development set \cite{yu2018spider}.

\begin{table*}
  \caption{Examples of inputs of our T5-small models.}
  \label{tab:ourprompt}
  \begin{tabular}{p{0.08\textwidth}|p{0.15\textwidth}|p{0.77\textwidth}}
    \hline
    \multirow{8}{*}{Task 1}&\multirow{3}{*}{column\_lm} & On average how large is the population of the counties? Answer 1 for county id,  Answer 2 for county name,  Answer 3 for population,  Answer 4 for zip code, the answer should be Answer  \\
    \cline{2-3}
    &\multirow{5}{*}{selection\_lm} & How many singers do we have? Answer 1 for *, Answer 2 for COUNT(*), Answer 3 for column, Answer 4 for COUNT(column), Answer 5 for SUM(column), Answer 6 for AVG(column), Answer 7 for MIN(column), Answer 8 for MAX(column), the answer should be Answer \\
    \hline
    \multirow{21}{*}{Task 2}&\multirow{13}{*}{column\_lm} & What is the average hours across all projects? SELECT [column], Answer 1 for code,  Answer 2 for name, Answer 3 for hours, the answer should be Answer \\
    \cline{3-3}
    && Find the ids of all the order items whose product id is 11. WHERE [column], Answer 1 for order item id,  Answer 2 for product id,  Answer 3 for order id,  Answer 4 for order item status,  Answer 5 for order item details, the answer should be Answer  \\
    \cline{3-3}
    && Find the number of followers for each user. GROUP BY [column], Answer 1 for user id,  Answer 2 for follower id, the answer should be Answer \\
    \cline{3-3}
    && List all pilot names in ascending alphabetical order. ORDER BY [column], Answer 1 for pilot id,  Answer 2 for name, Answer 3 for age, the answer should be Answer  \\
    \cline{2-3}
    &\multirow{6}{*}{selection\_lm} & How many singers do we have? Answer 1 for *, Answer 2 for COUNT(*), Answer 3 for column, Answer 4 for DISTINCT column, Answer 5 for COUNT(column), Answer 6 for COUNT(DISTINCT column), Answer 7 for SUM(column), Answer 8 for AVG(column), Answer 9 for MIN(column), Answer 10 for MAX(column), the answer should be Answer \\
    \cline{2-3}
    &\multirow{2}{*}{where\_lm} & How many king beds are there? Answer 1 for empty, Answer 2 for WHERE, the answer should be Answer \\
  \hline
\end{tabular}
\end{table*}

\begin{table*}
  \caption{Examples of baseline inputs.}
  \label{tab:baselineprompt}
  \begin{tabular}{p{0.22\textwidth}|p{0.78\textwidth}}
    \hline
    \multirow{6}{*}{T5-small} & Please show the categories of the music festivals with count more than 1. database is music\_4. tables are artist, volume, music festival. columns in artist are artist id, artist, age, famous title, famous release date. columns in volume are volume id, volume issue, issue date, weeks on top, song, artist id. columns in music festival are id, music festival, date of ceremony, category, volume, result. \\
    \hline
    \multirow{12}{*}{\makecell[c]{T5-small + CFGs \\(column\_lm)}}& What is the average hours across all projects? SELECT [column], Answer 1 for ssn,  Answer 2 for name, ..., Answer 7 for project, the answer should be Answer \\
    \cline{2-2}
      & Find the ids of all the order items whose product id is 11. WHERE [column], Answer 1 for customer id,  Answer 2 for customer name,  ...,  Answer 27 for order item id, the answer should be Answer \\
    \cline{2-2}
     & Find the number of followers for each user. GROUP BY [column], Answer 1 for user id,  Answer 2 for follower id,  ...,  Answer 11 for followers, the answer should be Answer \\
    \cline{2-2}
    & List all pilot names in ascending alphabetical order. ORDER BY [column], Answer 1 for pilot id, Answer 2 for name,  ...,  Answer 28 for aircraft id, the answer should be Answer \\
  \hline
\end{tabular}
\end{table*}

\end{document}